\documentclass[11pt]{article}
\usepackage{authblk}

\usepackage{amsmath,graphicx}
\usepackage[round,numbers, longnamesfirst,sort&compress]{natbib}
\usepackage{epstopdf}
\usepackage{marginnote}
\usepackage[margin=1in]{geometry}
\usepackage{pseudocode}
\usepackage{rotating}

\makeatletter
\def\tagform@#1{\maketag@@@{[#1]\@@italiccorr}}
\makeatother
\usepackage{setspace}
\usepackage{xr}
\usepackage{amssymb}
\usepackage{epstopdf}
\usepackage{subfigure}
\usepackage{color}
\usepackage{amsthm}
\usepackage{caption}

\title{{ \bf \large \center
Free-breathing cardiac MRI using bandlimited manifold modelling}}
\author{\it \small Sunrita Poddar$^*$}
\author{\it \small Yasir Mohsin$^*$}
\author{\it \small Deidra Ansah$^\dag$}
\author{\it \small Bijoy Thattaliyath$^\dag$}
\author{\it \small Ravi Ashwath$^\dag$}
\author{\it \small Mathews Jacob$^*$}
\affil{
\it \small * Department of Electrical and Computer Engineering, The University of Iowa, Iowa\\
\it \small \dag Pediatric Cardiology, University of Iowa, Iowa\\
}

\begin{document}
\maketitle
\hspace{-7mm}
Correspondence to:
Mathews Jacob\\
Department of Electrical and Computer Engineering\\
4016 Seamans Center\\
University of Iowa, IA 52242\\
Email: mathews-jacob@uiowa.edu\\

\noindent This work is supported by NIH 1R01EB019961-01A1.\\
\noindent Approximate word count : 4840\\
\noindent Number of figures \& tables: 8\
\newpage
\begin{abstract}
\noindent \textbf{Purpose:}
We introduce a novel bandlimited manifold framework and an algorithm to recover free-breathing and ungated cardiac MR images from highly undersampled measurements.\\
\noindent \textbf{Methods:}
The image frames in the free breathing and ungated dataset are assumed to be points on a bandlimited manifold. We introduce a novel kernel low-rank algorithm to estimate the manifold structure (Laplacian) from a navigator-based acquisition scheme. The structure of the manifold is then used to recover the images from highly undersampled measurements. A computationally efficient algorithm, which relies on the bandlimited approximation of the Laplacian matrix, is used to recover the images. 

\noindent \textbf{Results:}  The proposed scheme is demonstrated on several patients with different breathing patterns and cardiac rates, without requiring the need for manually tuning the reconstruction parameters in each case.\\
\noindent \textbf{Conclusion:}  The proposed scheme enabled the recovery of free-breathing and ungated data, providing reconstructions that are qualitatively similar to breath-held scans performed on the same patients. This shows the potential of the technique as a clinical protocol for free-breathing cardiac scans.
\\
\textbf{Key words: cardiac reconstruction, free-breathing, kernel methods, manifold models.}
\end{abstract}
\newpage	
\section*{INTRODUCTION}
\par Breath-held cine-MRI is the key component of a cardiac MRI exam, which offers valuable assessments about the structure and function of the heart. However, the acquisition of the data at high spatial and temporal resolution requires long breath-hold durations, which is often challenging for patients with chronic obstructive pulmonary disease (COPD) or obesity \cite{copd}. Pediatric patients who are unable to follow complex breath-holding instructions, often have to be sedated for the purpose of the scan. In addition, multiple breath-holds along with intermittent pauses for recovery also results in prolonged scan time, adversely impacting patient comfort and compliance. Several acquisition and reconstruction techniques have been introduced to improve cardiac cine MRI. Early work relied on the reduction of breath-hold durations in cine MRI by acquiring undersampled k-space measurements. The images were reconstructed by exploiting the structure of x-f space \cite{dime,paradise,blast}, diversity of coil sensitivities \cite{sense}, and the sparsity of k-space \cite{sparse}. Real-time methods that rely on parallel MRI \cite{blast} were introduced for subjects who cannot hold their breath, but these often suffer from lower image quality. Low-rank based schemes that rely on k-space navigators need different subspace/rank models for cardiac and non-cardiac spatial regions \cite{christodoulou, brinegar}, requiring user intervention. Another strategy is to estimate the cardiac and respiratory phases, and explicitly bin the data to their respective phases, followed by recovery using compressed sensing \cite{grasp,xdgrasp}. These schemes rely on a series of complex steps, including bandpass filtering using prior information about the cardiac and respiratory rates, and peak identification to estimate the phases.

We had recently introduced the SToRM \cite{storm} framework, which enables ungated cardiac cine imaging in the free-breathing mode using  radial acquisitions. The SToRM algorithm assumes that the images in the free-breathing dataset lie on a smooth and low-dimensional manifold, parameterized by a few variables (e.g. cardiac \& respiratory phases). The acquisition scheme relies on navigator radial spokes, which are used to compute the graph Laplacian matrix that captures the structure of the manifold. An off-diagonal entry of the Laplacian matrix is high if the corresponding pair of frames have similar cardiac and respiratory phases, even though they may be well-separated in time. This implicit soft-binning strategy offers the potential to simultaneously image cardiac and respiratory function, and eliminates the need for explicit binning of data as in \cite{grasp,xdgrasp}. Since the framework does not require the associated complex processing steps that assume the periodicity of the cardiac/respiratory motion, it is readily applicable to several dynamic applications, including speech imaging as shown in \cite{storm}, or cardiac applications involving arrhythmia. Conceptually similar manifold models have been proposed by other groups \cite{bhatia,usman}. Despite promising results, the above manifold models still have some deficiencies that restrict its clinical use. Specifically, the need to reconstruct and store the entire dataset (around 1000 frames) makes the algorithms memory demanding and computationally expensive, and restricts their eventual extension to 3-D applications. Another challenge that impairs the quality of the reconstruction is the sensitivity of the Laplacian estimation process to noise as well as subtle patient motion. 

In this work, we propose a bandlimited SToRM (b-SToRM) framework to overcome both of these challenges, and determine its utility on cardiac MRI patients. We introduce a bandlimited model for the manifold shape to improve the estimation of the Laplacian from the navigators. Specifically, we model the manifold in high dimensional  (equal to the number of pixels) space as the zero level-set of a band-limited potential function. We show that under the bandlimited assumption, exponential feature maps of each of the images can be annihilated by a finite impulse response filter whose support is the same as that of the Fourier co-efficients of the potential function. These annihilation relations translate to a low rank structure of the matrix of feature maps. We pose the recovery of the navigators from their noisy measurements as a nuclear norm minimization of the matrix of feature maps. We obtain a Laplacian matrix that is more robust to noise and subtle motion artifacts than the previous SToRM approach, as a by-product of the above optimization scheme. 

In order to reduce the computational complexity and memory demand by an order of magnitude, we approximate the Laplacian matrix by a few of its eigen vectors. The eigen vectors of the Laplacian are termed as Fourier exponentials on the manifold/graph \cite{ortegaGraph}. Instead of reconstructing the entire dataset, we propose to only recover the coefficients of the Laplacian basis functions. Since the proposed scheme in this paper improves the SToRM framework by using the bandlimited models, we refer to the new approach as the bandlimited SToRM (b-SToRM) framework. 

We validate the b-STORM framework on nine adult congenital heart disease patients with different imaging views, as an add-on to the routine contrast enhanced cardiac MRI study. We also study the impact of patient motion, reduced number of navigators, and reduced acquisition time on the algorithm. We show that the reconstructed images can be sorted into respiratory and cardiac phases using the eigen-vectors of the estimated Laplacian matrix, facilitating the easy visualization of the data. 

This work has similarities to the kernel low-rank approach for MRI reconstruction in \cite{nakarmi}. The algorithm in \cite{nakarmi} requires the computation of the feature maps of the polynomial kernel and their pre-images. The explicit computation of the feature maps is infeasible for Gaussian kernels since the feature maps are infinite-dimensional. Our approach relies on the kernel trick \cite{scholkopf}, and is thus computationally feasible for Gaussian kernels. This approach is built upon our recent work on annihilation based image recovery \cite{gregtsp,gregpapers,ongie2017fast} and the work on polynomial kernels introduced in \cite{gregvariety}; we extend \cite{gregvariety} to Gaussian kernels in this paper. 

\section*{THEORY}

\subsection*{Background on smooth manifold models for images}

A manifold is a topological space that locally resembles a Euclidean space. In particular, each $n$-dimensional point (where $n$ is the ambient dimension) on an $m$ dimensional manifold ($m<<n$) has a local neighbourhood, which has a continuous one-to-one mapping (homeomorphic) with a Euclidean space of dimension $m$. Many classes of natural images can be modelled as points sampled from a low-dimensional manifold, embedded in an ambient high dimensional space. The dimension of the ambient space is equal to the number of pixels in the images, while the dimension of the manifold depends on the degrees of freedom of the class of images. For example, a dataset of  images of faces of the same person may be parameterized  by  pose and lighting. Similarly, each image in a real-time cardiac MRI acquisition can parameterized by the cardiac and respiratory phases. 

Manifold embedding schemes \cite{lle,lapEig} aim to find a non-linear mapping  $f: \mathbb R^n \rightarrow \mathbb R^m$ from the points $\mathbf x_i \in \mathcal M \subset \mathbb R^n$, such that $f(\mathbf x_i) = \mathbf f_i$ preserves geodesic distances on the manifold; i.e, $\|\mathbf f_i-\mathbf f_j\|^2 \approx \|\mathbf x_i-\mathbf x_j\|_{\mathcal M}^2; \forall i,j$. As shown in \cite{lapEig}, in order to preserve local neighbourhoods of the manifold, we need function with low average smoothness  $\int_{\mathcal M} \|\nabla f\|^2$. Many algorithms, including the popular Laplacian eigen maps embedding algorithm \cite{lapEig} and the SToRM formulation \cite{storm} operate on discrete samples from the manifold. In terms of $k$ points $\{\mathbf x_i\} \in \mathbb R^n$, $i=1,\ldots,k$ sampled from the manifold, the average smoothness of $f$ can be approximated as: 
\begin{equation}
\label{unidimensional}
\int_{\mathcal M} \|\nabla f\|^2 \approx \frac{1}{2}\;\sum_{i,j=1}^k \mathbf W_{i,j}\; \|f(\mathbf x_i) -f(\mathbf x_j)\|^2 = {\rm trace}(\mathbf f\; \mathbf L\; \mathbf f^H)
\end{equation}
where the weight matrix $\mathbf W$ is specified by:
\begin{equation}
\label{Wcomp}
\mathbf W_{ij} = \begin{cases}
\mathbf e^{-\frac{d_{i,j}^{2}}{\sigma^{2}}}&, \text{if $\mathbf x_{i}$ and $\mathbf x_{j}$ are neighbours}.\\
0 &, \text{otherwise}.
\end{cases}
\end{equation}
Here, $d_{i,j}^2 = \|\mathbf x_i - \mathbf x_j\|^2$. The use of the exponential kernel assigns higher weights to local neighbours on the manifold. $\mathbf f$ is a matrix, whose columns correspond to $f(\mathbf x_i); i=1,..,k$ and $\mathbf L = \mathbf D - \mathbf W$ is the graph Laplacian, which approximates the Laplace Beltrami operator. $\mathbf D$ is a diagonal matrix with elements defined as $\mathbf D_{ii} = \sum_j \mathbf W_{ij}$. Three  approaches are used in the literature to determine if $\mathbf x_i$ and $\mathbf x_j$ are neighbours:
\vspace{-1.5em}
\begin{enumerate}
\item Distance thresholding: $\mathbf x_i$ and $\mathbf x_j$ are neighbours if $d_{ij} < t$, where $t$ is a fixed threshold. This may result in disconnected graphs.
\item Number of neighbours: $\mathbf x_i$ has only a fixed number of neighbours, which are the points with lowest distance from it. This technique always leads to fully connected graphs but may be associated with false edges.
\end{enumerate}
The optimal embedding obtained by minimizing \eqref{unidimensional} is a matrix of the eigen-vectors with the $m$ lowest eigen-values in the generalized eigen-vector problem $\mathbf L \mathbf f = \lambda \mathbf D \mathbf f$. 

\subsection*{SToRM framework \cite{storm}}

SToRM relies on a navigated radial acquisition scheme where the same navigator lines (2-4 radial spokes) are played at the beginning of every 10-12 spokes. The acquisition of the $i^{\rm th}$ frame can be represented as:
\begin{equation}
\label{measurements}
 \underbrace{\left[\begin{array}{c}
\mathbf z_{i,j}\\
\mathbf y_{i,j}
\end{array}
\right] }_{\mathbf b_{i,j}} =   \underbrace{\left[\begin{array}{c}
\boldsymbol\Phi\\
\mathbf B_{i}
\end{array}
\right] \mathbf F ~\mathbf C_j}_{\mathbf A_{ij}} ~\mathbf x_i + \boldsymbol\eta_{ij}
\end{equation}
Here, $\mathbf F$ is the 2-D Fourier transform matrix and $\mathbf C_j$ is a diagonal matrix corresponding to weighting by the $j^{\rm th}$ coil sensitivity map. $\boldsymbol \Phi$ is the sampling matrix corresponding to the navigators that is kept the same for all frames. 

The weight matrix of the image manifold is estimated from the navigators $\mathbf z_{i,l}=\boldsymbol\Phi\mathbf F ~\mathbf C_l ~\mathbf x_i, ~l=1,..,{\rm N_{coils}}$ using \eqref{Wcomp}, where:
 \begin{equation}
d_{ij}^2 = \sum_{l=1}^{\rm N_{coils}}\left\|\mathbf z_{il}-\mathbf z_{jl}\right\|^{2}
\end{equation}
This results in high weights between images with similar cardiac and respiratory phase, while the weights between  images with different phases are small. The manifold Laplacian $\mathbf L$ is computed from the weight matrix $\mathbf W$. 
 
The acquisition can be compactly rewritten as:
\begin{equation}
	\label{acquisition}
\mathbf B = \mathcal A(\mathbf X) + \boldsymbol{\eta}
\end{equation}
Here, $\mathbf X = \left[\mathbf x_1,\ldots,\mathbf x_k\right]$ is the Casorati matrix obtained by stacking the vectorized images as columns, while $\mathcal A$ captures the measurement process described in \eqref{measurements}. SToRM reconstructs the images by solving the following problem:
\begin{equation}
\mathbf X^{*} =\arg \min_{\mathbf X} \|\mathcal A(\mathbf X)-\mathbf B\|^{2}_{F} + \lambda~ \mathrm{{\rm trace}}(\mathbf X {\mathbf L} \mathbf X^{H})
\label{l2problem}
\end{equation}

A key drawback of SToRM is the sensitivity of the Laplacian matrix to noise and artifacts in the acquisition process. While the exponential weight choice is popular, this approach is is dependent on the specific way in which neighbours are selected. Another challenge is the large memory demand and computational complexity associated with the recovery of the large dataset, often consisting of 1000 frames. This makes it difficult to extend SToRM to 3D+time applications. 

The main focus of this paper is to introduce the b-SToRM framework, which minimizes the problems associated with SToRM \cite{storm}. We introduce a bandlimited manifold model for the systematic estimation of the Laplacian matrix from navigators. This reduces the sensitivity of the estimation process to noise and subtle patient motion. We also introduce an approximation of the Laplacian matrix using its eigen decomposition to  drastically reduce memory demand and computational complexity. The estimated Laplacian matrix enables the visualization of the reconstruction results. An overview of the proposed scheme is given in Fig 1.

\subsection*{Bandlimited manifold shape model \& kernel low-rank relation}
We model the manifold $\mathcal M \subset \mathbb R^n$ as the zero-level set of a bandlimited function $\psi: \mathbb R^n \rightarrow \mathbb R$, represented using its Fourier series (see Fig 2) :
\begin{equation}
\label{implicit}
\psi(\mathbf x) = \sum_{\mathbf k \in \Lambda} \mathbf c_{\mathbf k} e^{j~2\pi \mathbf k^T \mathbf x}
\end{equation}
The manifold is specified by the set of points $\{\mathbf x \in \mathbb R^n|\psi(\mathbf x)=0\}$. Here, $\Lambda \subset \mathbb Z^n$ is a set of contiguous discrete locations that indicate the support of the Fourier series co-efficients of $\psi$. We assume that $\{\mathbf c_k\}$ is the smallest set of Fourier co-efficients that satisfies the above relation; we term it as the minimal filter. We refer to the above representation as a bandlimited manifold. All points $\mathbf x$ on the implicit surface \eqref{implicit} satisfy $\psi(\mathbf x)=0$, which implies that: 
\begin{equation}
\label{nonlinearmapping}
\psi(\mathbf x) = \mathbf c^T \underbrace{ \begin{bmatrix} e^{j2\pi\mathbf k_1^T\mathbf x}\\ \vdots\\ e^{j2\pi\mathbf k_{|\Lambda|}^T\mathbf x}\end{bmatrix}}_{\phi_\Lambda (\mathbf x)}  = 0
\end{equation}
The entries of $\phi_\Lambda(\mathbf x)$ are non-linear transformations of $\mathbf x$, similar to kernel approaches \cite{scholkopf}; we term $\phi_\Lambda(\mathbf x)$ as the non-linear feature map of $\mathbf x$ (see Fig 2).  When there are multiple points $\mathbf x_1,.., \mathbf x_k$ sampled from the manifold, we have the following annihilation relation:
\begin{equation}
\label{matrixannihilation}
\mathbf c^T \underbrace{\begin{bmatrix}
\phi_{\Lambda}(\mathbf x_1),\ldots \phi_{\Lambda}(\mathbf x_k)
\end{bmatrix}
}_{\Phi_{\Lambda}(\mathbf X)} = 0
\end{equation}
Since $\mathbf c$ is the unique minimal filter of $\Phi_{\Lambda}(\mathbf X)$, rank$(\Phi_{\Lambda}(\mathbf X)) = |\Lambda|-1$. In practice, the exact bandwidth of $\Lambda$ is unknown. We choose a rectangular support $\Gamma \subset \mathbb Z^n$ such that $\Lambda \supseteq \Gamma$; the corresponding feature matrix is denoted by $\Phi_{\Gamma}(\mathbf X)$. $\mathbf c_1$ obtained by zero-padding the original coefficients $\mathbf c$ will satisfy $\mathbf c_1^T \Phi_{\Gamma}(\mathbf X) = 0$. $\mathbf c_2$ obtained by shifting $\mathbf c_1$ by an integer value will also satisfy $\mathbf c_2^T \Phi_{\Gamma}(\mathbf X) = 0$. We denote the number of valid shifts of $\mathbf c$ such that it is still support limited in $\Gamma$ by $|\Gamma:\Lambda|$ \cite{gregpapers}. Thus, we have: 
\begin{equation}
{\rm rank}(\Phi_{\Gamma}(\mathbf X)) \leq |\Gamma|-|\Gamma:\Lambda|
\end{equation}
If the number of points $k$ is greater than the above rank, we obtain right null-space relations $\Phi_{\Gamma}(\mathbf X)~\mathbf v_i = \mathbf 0$, or equivalently, $\mathbf K^{\Gamma}\mathbf v_i = \mathbf 0$. The entries of the $P\times P$ Gram matrix $\mathbf K^{\Gamma} = \Phi_{\Gamma}(\mathbf X)^H\Phi_{\Gamma}(\mathbf X)$ are given by: 
\begin{equation}
\mathbf K_{i,j}^{\Gamma} = \phi_{\Gamma}(\mathbf x_i)^H\phi_{\Gamma}(\mathbf x_j) =\underbrace{ \sum_{k\in \Gamma}  e^{\left(j~2\pi\mathbf k^T \left(\mathbf x_j-\mathbf x_i\right)\right)}}_{\kappa_{\Gamma}(\mathbf x_j-\mathbf x_i)}
\end{equation}
where $\kappa_{\Gamma}(\mathbf r)$ is shift invariant. When $\Gamma$ is a centered cube in $\mathbf R^n$, we have the Dirichlet kernel $\kappa_{\Gamma}(\mathbf r) = {\rm D}_{\Gamma}(\mathbf r)$. The above relations show that when the points live on the level-sets of a bandlimited function, their Dirichlet kernel matrix is low-rank. If we choose the weighted maps $\phi'_\Gamma = \mathbf M ~\phi_\Gamma$ (where $\mathbf M$ is a diagonal matrix with entries $e^{-\pi^2 \sigma^2 \mathbf \|\mathbf k\|^2}$), then the kernel function approaches a Gaussian as $\Gamma \rightarrow \mathbb Z^n$. In this case, the matrix $\mathbf K_{\Gamma}$ is theoretically full rank. However, we observe that the Fourier series coefficients of a Gaussian function can be safely approximated to be zero outside $|\mathbf k|< 3/\pi\sigma$, which translates to $|\Lambda| \approx \left(\frac{6}{\pi\sigma}\right)^n$; i.e., the rank will be small for high values of $\sigma$. 

The above results show that the rank of the feature map matrix $\Phi_\Gamma$ or the kernel matrix $\mathbf K_{\Gamma}$ can be used as a measure of the smoothness of the manifold. Specifically, if the rank is small, a low bandwidth implicit surface $\psi$ is sufficient to annihilate all the images in the dataset; this implies that the points lie on a smooth manifold, which is the zero level set of a bandlimited $\psi$.  

\subsection*{Laplacian estimation using feature-map rank minimization}
\label{lapest}
We use this low-rank prior to estimate the Laplacian matrix of the manifold. Since this approach is more systematic compared to the exponential weight based technique, it is expected to be more robust to noise and related artifacts. We recover the navigator signals $\mathbf R$ from their noisy measurements $\mathbf Z$ by solving the following optimization problem:
\begin{equation}
\label{kernel}
\mathbf R^*  = \arg\min_{\mathbf R}  \|\mathbf R - \mathbf Z\|_F^2 + \mu~
\left\| \Phi\left(\mathbf R\right)\right\|_*
\end{equation}
where $\|\cdot\|_*$ denotes the nuclear norm and $\Phi(\mathbf R)$ denotes a matrix whose columns are the non-linear maps of the columns of $\mathbf R$ (corresponding to different frames), similar to \eqref{nonlinearmapping}. Note that the above formulation simplifies to low-rank denoising similar to \cite{ktslr} when $\Phi = \mathcal I$, which is the identity map. 

Inspired by similar methods in low-rank recovery \cite{irls}, we introduce an iterative reweighted least squares (IRLS) algorithm  to solve the above minimization problem. The approach followed is similar to \cite{gregvariety}, where the case of polynomial kernels is discussed. The manifold Laplacian is obtained as a byproduct of this algorithm. This approach relies on approximating the nuclear norm penalty in \eqref{kernel} as:
\begin{equation}
\left\|\Phi(\mathbf R)\right\|_* = {\rm trace}\left[\left(\Phi(\mathbf R)^T \Phi(\mathbf R)\right)^{\frac{1}{2}}\right] = {\rm trace}\left[\mathcal K(\mathbf R)^{\frac{1}{2}}\right] \approx {\rm trace}\left[\mathcal K(\mathbf R)\mathbf P\right]
\end{equation}
where $\mathbf P = \left[\mathcal K(\mathbf R) + \gamma \mathbf I\right]^{-\frac{1}{2}}$ and $\mathcal K$ is the Gaussian kernel. The algorithm alternates between the following 2 steps: 
\begin{equation}
\label{kernelApprox}
\mathbf R^{(n)}= \arg\min_{\mathbf R} \underbrace{ \|\mathbf R - \mathbf Z\|_F^2 + \mu~
{\rm trace}\left[\mathcal K(\mathbf R)\mathbf P^{(n)}\right]}_{\mathcal C(\mathbf R^{(n)})}
\end{equation}
\begin{equation}
\mathbf P^{(n)} = \left[\mathcal K\left(\mathbf R^{(n-1)}\right) + \gamma^{(n)} \mathbf I\right]^{-\frac{1}{2}}
\end{equation}
where $\gamma^{(n)} = \frac{\gamma^{(n-1)}}{\eta}$, and $\eta>1$ is a constant.
 
We use the kernel trick $\left\langle \Phi(\mathbf r_1),\Phi(\mathbf r_2)\right\rangle = \mathcal K\left(\mathbf r_1,\mathbf r_2\right)$ to solve \eqref{kernel} without explicitly evaluating the maps $\Phi(\mathbf R)$. Note that $\Phi(\mathbf R)$ may be considerably higher in dimension compared to the frames. A key benefit of this approach over the kernel low-rank approach in \cite{nakarmi} is that we do not require the computation of explicit images and pre-images, and hence our scheme is applicable to shift invariant kernels that are widely used. 

The estimation of $\mathbf R$ involves solving a non-linear system of equations. To reduce computational complexity,  we linearize the gradient of the cost function in \eqref{kernelApprox} with respect to $\mathbf R$. The gradient of the objective function w.r.t $\mathbf R_i$ is given by $ \nabla_{\mathbf R_i}\mathcal C = 2(\mathbf R_i - \mathbf Z_i) + \mu \sum_j \nabla_{\mathbf R_i}[\mathcal K(\mathbf R)]_{ij}\mathbf P^{(n)}_{ij}$. Assuming a Gaussian kernel and linearizing the gradient with respect to $\mathbf R_i$, we obtain:
\begin{equation}
\begin{split}
\nabla_{\mathbf R_i} \mathcal C& \approx 2(\mathbf R_i - \mathbf Z_i) +2\mu \sum_j w_{ij}^{(n-1)}(\mathbf R_i- \mathbf R_j)
\end{split}
\end{equation}
where $w_{ij}^{(n-1)}$ is the $(i,j)^{th}$ entry of a matrix $\mathbf W^{(n-1)} = -\frac{1}{\sigma^2}\mathcal K(\mathbf R^{(n-1)}) \odot \mathbf P^{(n)}$. In matrix form, the gradient can be rewritten as $\nabla_{\mathbf R} \mathcal C  = 2(\mathbf R - \mathbf Z) +  2\mu \mathbf R \mathbf L^{(n-1)}$, 
where $\mathbf L^{(n)}$ is the Laplacian matrix computed from the weight matrix $\mathbf W^{(n)}$. This results in the following equivalent optimization problem for the estimation of $\mathbf R$ at the $n^{th}$ iteration, which can be solved analytically:
\begin{equation}
\label{Req}
\mathbf R^{(n)}  = \arg\min_{\mathbf R} \left\|\mathbf R - \mathbf Z\right\|_F^2 + \mu~{\rm trace}\left(\mathbf R~ \mathbf L^{(n)} ~\mathbf R^{H} \right)
\end{equation}
Note that the above iterative algorithm is analogous to SToRM, where $\mathbf L^{(n)}$ is the Laplacian. We use $\mathbf L^{(n)}$ obtained from the above denoising problem to recover the image frames from their undersampled measurements.

\subsection*{Efficient signal recovery using bandlimited approximation of the Laplacian matrix}
\label{secondstep}
The SToRM implementation \eqref{l2problem} required the storage and processing of a large number of frames (around 400-1000), which is computationally expensive. We now propose an efficient algorithm based on the eigen decomposition of the Laplacian matrix to significantly reduce the computational complexity and memory demand. 
Denoting the eigen decomposition of the symmetric Laplacian matrix as $\mathbf L = \mathbf V\Sigma \mathbf V^H$, we rewrite the SToRM cost function in \eqref{l2problem} as: 
\begin{eqnarray}
\mathbf X^{*} &=& \arg \min_{\mathbf X} \|\mathcal A(\mathbf X)-\mathbf B\|^{2}_{F} + \lambda~ \mathrm{{\rm trace}}\left[\underbrace{(\mathbf X \mathbf V)}_{\mathbf U}~{\boldsymbol \Sigma} \underbrace{(\mathbf X \mathbf V)^{H}}_{\mathbf U^H})\right]\\
&=& \arg \min_{\mathbf X} \|\mathcal A(\mathbf X)-\mathbf B\|^{2}_{F} + \lambda~ \sum_{i=1}^k \sigma_i\left\|\underbrace{\mathbf X\,\mathbf v_i}_{\mathbf u_i}\right\|^2
\end{eqnarray}
Here, the columns of $\mathbf V$ form an orthonormal temporal basis set and $\mathbf u_i$ are the spatial coefficients. We observe that the minimum eigen value of the Laplacian matrix is zero, while the other eigen values often increase rapidly. Hence, the weighted norm in the penalty encourages signals $\mathbf X$ that are maximally concentrated along the eigen vectors $\mathbf v_i$ with small eigen values; these eigen vectors correspond to smooth signals on the manifold. Since the projections of the recovered signal onto the higher singular vectors are expected to be small, we pick the $r$ smallest eigen vectors of $\mathbf L$ to approximate the recovered matrix as: 
\begin{equation}
\mathbf X = \mathbf U_{r} \mathbf V_{r}^H
\end{equation}
where $\mathbf U_r$ is a matrix of $r$ basis images (typically around $r \approx 30$) and $\mathbf V_r$ is a matrix of $r$ eigen vectors of $L$ with the smallest eigen values. Thus the optimization problem \eqref{l2problem} now reduces to:
\begin{equation}
\mathbf U^{*} =\arg \min_{\mathbf U} \|\mathcal A(\mathbf U\mathbf V^H)-\mathbf B\|^{2}_{F} + \lambda~ \sum_{i=1}^r \sigma_i \|\mathbf u_i\|^2
\label{l2synthesis}
\end{equation} 

We observe $r\approx 30$ is sufficient to recover the dynamic dataset with high accuracy. Typically, a dataset with less amount of motion can be accurately represented with a lower value of $r$. However, we choose a fixed larger value ($r=30$), which would work for the more challenging datasets as well. Note that in this case, the reconstruction algorithm aims to recover $r$ coefficient images from the measurement data; the optimization problem is expected to be an order of magnitude less computationally intensive than \eqref{l2problem}, especially when the number of basis functions $r$ is low. 

\subsection*{Visualization of the reconstructed data using manifold embedding}
Laplacian eigen-maps rely on the eigen vectors of the Laplacian matrix to embed the manifold to a lower dimensional space. When the signal variation in the dataset is primarily due to cardiac and respiratory motion, the second and third lowest eigen vectors are often representative of the cardiac and respiratory phases. This information may be used to bin the recovered data into respiratory and cardiac phases for visualization as in Fig 7, even though we do not use explicit binning for image recovery. This post-processing step can be thought of as a manifold embedding scheme using an improved Laplacian eigen-maps algorithm \cite{lapEig}, where the main difference with \cite{lapEig} is the estimation of the Laplacian. 

\section*{METHODS}

Cardiac data was collected in the free-breathing mode from nine patients at the University of Iowa Hospitals and Clinics on a 1.5 T Siemens Aera scanner. The institutional review board at the local institution approved all the in-vivo acquisitions and written consent was obtained from all subjects. A FLASH sequence was used to acquire 10 radial lines per frame out of which 4 were uniform radial navigator lines and 6 were Golden angle lines. The sequence parameters were: TR/TE=4.3/1.92 ms, FOV=300mm, Base resolution=256, Bandwidth=574Hz/pix. 10000 spokes of k-space were collected in 43 s. Data corresponding to two views (two-chamber/short-axis and four-chamber) was collected for each patient, resulting in a total of 18 datasets. We used the proposed scheme to reconstruct these datasets. The parameters of the image reconstruction algorithm were manually optimized on one dataset, and kept fixed for the rest of the datasets. 

We conduct a few experiments to study the performance of our method in different datasets, and with different acquisition parameters. We study the impact of motion patterns on the reconstructions, using two of the most challenging datasets, with different breathing and cardiac patterns. We also study the effect of the number of navigator lines on the quality of the recovered images using a dataset with a large amount of breathing motion. The main goal is to determine the minimum number of navigator lines per frame to acquire  in future studies. For this purpose, we compared the reconstruction using 4 navigator lines to that using only 1 or 2 navigator lines. Two experiments were conducted using 2 navigator lines per frame (corresponding to $0^\circ$ and $90^\circ$)  and 1 navigator line per frame (corresponding to $0^\circ)$  respectively to estimate the weights. For the purpose of reconstruction, we used the full data (6 golden angle lines and 4 navigators). We also study the impact of the acquisition duration on image quality. For 2 datasets with different types of motion patterns, we compare the reconstruction using the entire data, 450 contiguous frames corresponding to 22 s, and also 300 frames corresponding to 12 s.

We demonstrate that the recovered data can be automatically binned into respiratory and cardiac phases using two eigen-vectors of the estimated Laplacian matrix. Thanks to the accurate and robust estimation of the Laplacian matrix, these eigen-vectors accurately represent the respiratory and cardiac motion of the patient over the entire acquisition. Using this information, each image frame can be assigned a bin depending on its respiratory and cardiac phase. Images from each bin can be viewed to find representative members of a particular cardiac or respiratory phase. We also compare our free-breathing ungated reconstructions to images obtained from a clinical breath-held sequence on the same patients.

\section*{RESULTS}
The benefit of the low-rank Laplacian estimation scheme over the SToRM estimate, which is based on the exponential kernel \cite{lapEig}, is evident from Fig. 3. The navigator signals denoised using the proposed scheme show the preservation of the cardiac and respiratory motion, while reducing the impact of noise and subtle patient motion. The temporal basis functions (lowest eigen vectors of the Laplacian) when the Laplacian matrix is computed using (c) the proposed iterative scheme (d) the exponential scheme (e) the exponential scheme where only two neighbours have been retained per frame also show the benefit of the the proposed Laplacian estimation scheme. The new approach captures the cardiac and respiratory variations, while both other strategies are highly sensitive to abrupt patient motion. Since most of the information is captured by the lower eigen vectors as seen from Fig 3 (c), a bandlimited approximation of the Laplacian matrix is sufficient to solve for the dataset.

The datasets in Fig 4 have a high amount of respiratory and out-of plane motion, compared to the other datasets that we have collected. The first dataset shows a normal cardiac rate (68 beats/min) accompanied by a very irregular breathing pattern, characterized by several large gasps of breath. We show a few reconstructed frames from different time points, at various states of motion. The reconstruction quality is better in presence of less respiratory motion since there are frames similar to it in the dataset; the manifold neighbourhood is well sampled in these neighbourhoods. By contrast, the images are seen to be more noisy in manifold regions that are not well-sampled (red box). The second dataset shows a high cardiac rate (107 beats/min) accompanied by heavy regular breathing (42 breaths/min). We observe that the algorithm is able to reconstruct this case satisfactorily, despite the rapid motion since the manifold is well-sampled. 

We observe from Fig 5 that for both high and low motion regions, there is no degradation in image quality when the number of navigator lines are reduced to two from four. Only using one navigator spoke induces some error, especially for the frames highlighted in green since they have more respiratory motion. This is expected since the approach will only be sensitive to the motion in one direction and not to the direction orthogonal to it. As a result of this experiment, we plan to keep only two navigator lines per frame in the future, and consequently increase the number of golden angle lines to 8 (from 6 in the current acquisition). This should improve image quality by making the sampling patterns between frames more incoherent.

The effect of reducing acquisition time is illustrated in Fig 6. The dataset at the top has more breathing motion as compared to the  bottom one. We observe that the bottom dataset is robust to decrease in the number of frames; it can be reliably recovered even from 12 seconds of data. The top dataset is more sensitive to reduction in scan time. The green line corresponds to the lowest position of the diaphragm, which is less frequent in the dataset. By contrast, the blue line corresponds to a more frequent frame. The frames around  the green line, shown in the green box are more noisy when the scan time is reduced to 12 seconds, compared to the reconstructions within the blue box. We observe negligible errors in both datasets when the acquisition time is reduced to 22s, whereas relatively noisy reconstructions are seen in high motion frames when it is reduced to 12 second acquisition windows. The error images for Fig 5 and Fig 6 are on the same scale, to illustrate the relative effects of changing the number of navigators and the number of frames. 

The results in Fig 7 show that the improved Laplacian eigen maps approach facilitates the easy visualization of the data. In general, we observe that the eigen-vectors of the Laplacian matrix with the second and third lowest eigen values correspond to respiratory and cardiac motion. It can be appreciated from Fig 3 that such a binning strategy is not possible when the exponential weights are used. 

Fig 8 demonstrates the potential of our proposed scheme to replace clinical breath-held and gated techniques. There is some difference in the appearance of the breath-held and free-breathing reconstructions due to mismatch in slice position. Moreover, the breath-held acquisition was done using a TRUFI sequence, and thus shows higher contrast than the free-breathing data which was acquired using a FLASH sequence. In spite of these differences, we note that the images reconstructed using our proposed scheme are of clinically acceptable quality.

\section*{DISCUSSION}
We have introduced the b-STORM framework for the recovery of free-breathing and ungated cardiac images in a short 2-D acquisition. We assume that the images are points on a smooth manifold. We estimate the graph Laplacian from radial navigators. This framework relies on two key innovations over the SToRM algorithm: \textbf{(i)} A novel algorithm imposing a bandlimited manifold model, is used to estimate the Laplacian matrix; the new estimate is considerably more robust to noise and subtle noise variation. \textbf{(ii)} A bandlimited approximation of the Laplacian reduces the computational complexity and memory demand of the algorithm by an order of magnitude.  Due to its computational efficiency and lack of need for manual intervention, the b-SToRM framework may be a good candidate for clinical scans where patients (e.g. pediatric patients, patients with COPD) are unable to hold their breath for sufficiently long periods of time, or are unable to follow breath-holding instructions. We plan to extend the proposed scheme to include perfusion and parameter mapping, thus moving towards a single short free-breathing clinical cardiac MR scan for structural and functional imaging. We also plan to extend the method for 3D acquisitions. 
  
While the framework has similarities to low-rank approaches \cite{ktslr,zhao_psf_2010,bcs} that rely on the factorization of the Casorati matrix, the key difference is the signal model and the approach in which the temporal basis functions are estimated. Conventional low-rank methods often require the binning of the k-space data to respiratory bins before reconstructions, using self gating approaches \cite{grasp}. The main benefit of the proposed scheme is that it does not require any explicit binning, which often includes complex steps including band-pass filtering, and peak isolation. The computational complexity and memory demand of the algorithm is comparable to XD-GRASP and similar binning approaches, thanks to the bandlimited Laplacian approximation. 

We demonstrate our algorithm on a number of datasets with different respiratory and cardiac patterns. In accordance with the results of our retrospective experiments on the impact of the number of navigator lines, we plan to collect data with only two navigator lines in the future. This would increase the incoherence of the undersampling patterns across frames, resulting in better quality reconstructions. Our experiments on reduced scan time shows that we can obtain reliable data from datasets with high motion with around 22s of data/slice, while it can be pushed down to 12s for datasets with less motion. We have not imposed any spatial regularization on the recovered images. Perhaps, with the addition of such priors, the acquisition time can be further reduced.

Our method produces a series of ungated images, enabling the user to visualize the real-time data with both  respiratory and cardiac motion. This approach may be useful in studies on patients with pulmonary complications such as COPD. The data can also be automatically segmented into respiratory and cardiac phases post reconstruction for easy visualization of the data, using the eigen-vectors of the estimated Laplacian matrix. 

Since the study was an add on to the routine cardiac exam, there was no perfect control on the specific time point of acquisition following contrast administration. This explains the differing contrast between the datasets. 

\section*{CONCLUSION}
We proposed a novel bandlimited manifold regularization framework termed as b-SToRM for free-breathing and ungated cardiac MR imaging. The validation of the dataset using cardiac datasets with differing amount of cardiac and respiratory motion shows the ability of the scheme to provide good image quality. It is also demonstrated that the resulting ungated images can be easily binned into respiratory and cardiac phases and viewed as a gated dataset. The success of the method on very challenging datasets with high cardiac rate and irregular breathing patterns suggests a useful clinical application of the method on patients who have difficulty in following traditional breath-holding instructions. 

\newpage
\section*{Legends}
\textbf{Fig 1:} {\small Outline of the b-SToRM scheme. The free breathing and ungated data is acquired using a navigated golden angle acquisition scheme. We estimate the Laplacian matrix from the navigator data using the kernel low-rank model. The entries of the Laplacian matrix specifies the connectivity of the points of the manifold, with larger weights between similar frames in the dataset. The manifold is illustrated by the sphere, while the connectivity of the points are denoted by lines whose thickness is indicative of the proximity on the manifold. Note that the frames that may be closer on the manifold may be well separated in time. The bandlimited manifold recovery scheme uses the Laplacian matrix to recover the images from the acquired k-space measurements. The Laplacian matrix also facilitates the easy visualization of the data.} \\

\textbf{Fig 2:} {\small Illustration of Annihilation Condition: The data points $\mathbf x$ lie on the zero-level set of a band-limited function $\psi$. Thus, each point $\mathbf x$ satisfies the relation: $\psi(\mathbf x) = 0$. The Fourier series co-efficients $\mathbf c$ satisfy the annihilation relation $\mathbf c^T \phi(\mathbf x) = 0$, where $\phi(\mathbf x)$ is a non-linear feature mapping of $\mathbf x$.} \\

\textbf{Fig 3:} {\small Improved manifold Laplacian estimation using iterative low-rank approach: We compare the proposed scheme against the old SToRM approach that relies on exponential kernels. The proposed scheme denoises the original navigator signals in (a) using the low-rank approach to obtain (b). It is seen that the denoising approach significantly reduces the noise, while retaining the cardiac and respiratory motion information. The eigen functions $\mathbf V$ with the smallest eigen values of  the Laplacian estimated using the proposed scheme, exponential weights, and exponential weights with truncation (SToRM approach) are shown in (c), (d), and (e), respectively. It is observed that the proposed scheme provides more accurate estimates of cardiac and respiratory motion  than the other schemes, facilitating the low-rank approximation. Both (d) and (e) are affected by subtle motion of the subject while the proposed scheme is relatively unperturbed. The spatial coefficients $\mathbf U$ estimated for each case is also shown.}\\

\textbf{Fig 4:} {\small Sensitivity of the algorithm to high motion: We illustrate the proposed scheme on two datasets acquired from two patients with different types of motion. For both datasets, we show a temporal profile for the whole acquisition to give an idea of the amount of breathing and cardiac motion present. We also show a few frames from time points with varying respiratory phase. The dataset on the left has regions with abrupt breathing motion at a few time points. Since these image frames have few similar frames in the dataset (poorly sampled neighbourhood on the manifold), the algorithm results in slightly noisy reconstructions at the time points with high breathing  motion (red box). The regions with low respiratory motion (blue and light blue boxes) are recovered well. The dataset on the right shows consistent, but low respiratory motion. By contrast, the heart rate in this patient was high. We observe that the proposed algorithm is able to produce good quality reconstructions in this case, since all neighbourhoods of the manifold are well sampled.} \\

\textbf{Fig 5:} {\small Effect of number of navigator lines on the reconstruction quality. We perform an experiment to study the effect of computing the Laplacian matrix $\mathbf L$ from different number of navigator lines. For this purpose, we use one of the acquired datasets with 4 navigator lines per frame. We compute the ground-truth $\mathbf L$ matrix using all 4 navigators. Next, we also estimate the $\mathbf L$ matrix using 2 navigator lines (keeping only the $0^\circ$ and $90^\circ$ lines) and 1 navigator line (keeping only the $0^\circ$ line). We now reconstruct the full data using these three Laplacian matrices, as shown in the figure. We observe that two navigator lines are sufficient to compute the Laplacian matrix reliably. Using one navigator line induces some errors, especially in the frames highlighted in green which are from a time point with higher respiratory motion. As a comparison, note that the error images are in the same scale as those for Fig 6.}\\

\textbf{Fig 6:} {\small Effect of number of frames on the reconstruction quality. We perform an experiment to study the effect of reconstructing the data from a fraction of the time-frames acquired. The original acquisition was 45 seconds long, resulting in 1000 frames. We compare the reconstruction of the $1^{st}$ 250 frames, using (1) all 1000 frames (2) only 550 frames, i.e. 22 s of acquisition (3) only 350 frames, i.e. 12 s of acquisition. As can be seen from the temporal profiles, Dataset-1 has more respiratory motion than Dataset-2. Consequently, the performance degradation in Dataset-1 is more pronounced with decrease in the number of frames. Moreover, the errors due to decrease in the number of frames is mostly seen in frames with higher respiratory motion, as pointed out by the arrows. As a comparison, note that the error images are in the same scale as those for Fig 5. }\\

\textbf{Fig 7:} {\small Binning into cardiac and respiratory phases. We demonstrate that the reconstructed ungated image series can easily be converted to a gated series of images if desired. For this purpose, the $2^{nd}$ and $3^{rd}$ eigen-vectors of the estimated Laplacian matrix are used as an estimate of the respiratory and cardiac phases respectively. The images can then be separated into the desired number of cardiac and respiratory bins. Here, we demonstrate this on two datasets that have been separated into 8 cardiac and 4 respiratory phases. Representative images from these bins have been shown in the figure.}\\

\textbf{Fig 8:} {\small Comparison to breath-held scheme. We demonstrate that our proposed free-breathing reconstruction technique produces images of similar quality to clinical breath-held scans, in the same acquisition time. Note that there are differences between the free-breathing and breath-held images due to variations in contrast between TRUFI and FLASH acquisitions, and also due to mismatch in slice position. However, the images we obtain are of clinically acceptable quality. Moreover, unlike the breath-held scheme we reconstruct the whole image time series (as is evident from the temporal profile). This can provide richer information , such as studying the interplay of cardiac and respiratory motion.}\\

\newpage
\bibliographystyle{unsrtnat-mrm}
\bibliography{refs}

\begin{thebibliography}{26}
\providecommand{\natexlab}[1]{#1}
\providecommand{\url}[1]{\texttt{#1}}
\expandafter\ifx\csname urlstyle\endcsname\relax
  \providecommand{\doi}[1]{doi: #1}\else
  \providecommand{\doi}{doi: \begingroup \urlstyle{rm}\Url}\fi

\bibitem[Gay et~al.(1994)Gay, Sistrom, Holder, and Suratt]{copd}
Gay SB, Sistrom CL, Holder CA, and Suratt PM.
\newblock {B}reath-{H}olding {C}apability of {A}dults: {I}mplications for
  {S}piral {C}omputed {T}omography, {F}ast-{A}cquisition {M}agnetic {R}esonance
  {I}maging, and {A}ngiography.
\newblock {Investigative Radiology}, 1994; 29.

\bibitem[Liang et~al.(1997)Liang, Jiang, Hess, and Lauterbur]{dime}
Liang ZP, Jiang H, Hess CP, and Lauterbur PC.
\newblock Dynamic imaging by model estimation.
\newblock {International journal of imaging systems and technology}, 1997;
  8:\penalty0 551--557.

\bibitem[Sharif et~al.(2010)Sharif, Derbyshire, Faranesh, and
  Bresler]{paradise}
Sharif B, Derbyshire JA, Faranesh AZ, and Bresler Y.
\newblock Patient-adaptive reconstruction and acquisition in dynamic imaging
  with sensitivity encoding (paradise).
\newblock {Magnetic Resonance in Medicine}, 2010; 64:\penalty0 501--513.

\bibitem[Tsao et~al.(2003)Tsao, Boesiger, and Pruessmann]{blast}
Tsao J, Boesiger P, and Pruessmann KP.
\newblock k-t blast and k-t sense: Dynamic mri with high frame rate exploiting
  spatiotemporal correlations.
\newblock {Magnetic resonance in medicine}, 2003; 50:\penalty0 1031--1042.

\bibitem[Pruessmann et~al.(1999)Pruessmann, Weiger, Scheidegger, Boesiger,
  et~al.]{sense}
Pruessmann KP, Weiger M, Scheidegger MB, Boesiger P, and others .
\newblock Sense: sensitivity encoding for fast mri.
\newblock {Magnetic resonance in medicine}, 1999; 42:\penalty0 952--962.

\bibitem[Lustig et~al.(2006)Lustig, Santos, Donoho, and Pauly]{sparse}
Lustig M, Santos JM, Donoho DL, and Pauly JM.
\newblock kt sparse: High frame rate dynamic mri exploiting spatio-temporal
  sparsity.
\newblock In {Proceedings of the 13th Annual Meeting of ISMRM, Seattle}, 2006;
  volume 2420.

\bibitem[Christodoulou et~al.(2014)Christodoulou, Hitchens, Wu, Ho, and
  Liang]{christodoulou}
Christodoulou AG, Hitchens TK, Wu~YL, Ho~C, and Liang ZP.
\newblock Improved subspace estimation for low-rank model-based accelerated
  cardiac imaging.
\newblock {IEEE Transactions on Biomedical Engineering}, 2014; 61:\penalty0
  2451--2457.

\bibitem[Brinegar et~al.(2010)Brinegar, Schmitter, Mistry, Johnson, and
  Liang]{brinegar}
Brinegar C, Schmitter SS, Mistry NN, Johnson GA, and Liang ZP.
\newblock Improving temporal resolution of pulmonary perfusion imaging in rats
  using the partially separable functions model.
\newblock {Magnetic Resonance in Medicine}, 2010; 64:\penalty0 1162--1170.
\newblock ISSN 1522-2594.
\newblock \doi{10.1002/mrm.22500}.
\newblock URL \url{http://dx.doi.org/10.1002/mrm.22500}.

\bibitem[Feng et~al.(2014)Feng, Grimm, Block, Chandarana, Kim, Xu, Axel,
  Sodickson, and Otazo]{grasp}
Feng L, Grimm R, Block KT, Chandarana H, Kim S, Xu~J, Axel L, Sodickson DK, and
  Otazo R.
\newblock Golden-angle radial sparse parallel mri: Combination of compressed
  sensing, parallel imaging, and golden-angle radial sampling for fast and
  flexible dynamic volumetric mri.
\newblock {Magnetic resonance in medicine}, 2014; 72:\penalty0 707--717.

\bibitem[Feng et~al.(2016)Feng, Axel, Chandarana, Block, Sodickson, and
  Otazo]{xdgrasp}
Feng L, Axel L, Chandarana H, Block KT, Sodickson DK, and Otazo R.
\newblock Xd-grasp: Golden-angle radial mri with reconstruction of extra
  motion-state dimensions using compressed sensing.
\newblock {Magnetic resonance in medicine}, 2016; 75:\penalty0 775--788.

\bibitem[Poddar and Jacob(2016)]{storm}
Poddar S and Jacob M.
\newblock Dynamic mri using smoothness regularization on manifolds (storm).
\newblock {IEEE Tran. Medical Imaging}, April2016; 35:\penalty0 1106--1115.
\newblock ISSN 0278-0062.
\newblock \doi{10.1109/TMI.2015.2509245}.

\bibitem[Bhatia et~al.(2015)Bhatia, Caballero, Price, Sun, Hajnal, and
  Rueckert]{bhatia}
Bhatia KK, Caballero J, Price AN, Sun Y, Hajnal JV, and Rueckert D.
\newblock Fast reconstruction of accelerated dynamic mri using manifold kernel
  regression.
\newblock In {International Conference on Medical Image Computing and
  Computer-Assisted Intervention}, 2015; pages 510--518. Springer.

\bibitem[Usman et~al.(2015)Usman, Atkinson, Kolbitsch, Schaeffter, and
  Prieto]{usman}
Usman M, Atkinson D, Kolbitsch C, Schaeffter T, and Prieto C.
\newblock Manifold learning based ecg-free free-breathing cardiac cine mri.
\newblock {Journal of Magnetic Resonance Imaging}, 2015; 41:\penalty0
  1521--1527.

\bibitem[Ortega et~al.(2017)Ortega, Frossard, Kova{\v{c}}evi{\'c}, Moura, and
  Vandergheynst]{ortegaGraph}
Ortega A, Frossard P, Kova{\v{c}}evi{\'c} J, Moura JM, and Vandergheynst P.
\newblock Graph signal processing.
\newblock {arXiv preprint arXiv:1712.00468}, 2017.

\bibitem[Nakarmi et~al.(2017)Nakarmi, Wang, Lyu, Liang, and Ying]{nakarmi}
Nakarmi U, Wang Y, Lyu J, Liang D, and Ying L.
\newblock A kernel-based low-rank (klr) model for low-dimensional manifold
  recovery in highly accelerated dynamic mri.
\newblock {IEEE Transactions on Medical Imaging}, 2017; PP:\penalty0 1--1.
\newblock ISSN 0278-0062.
\newblock \doi{10.1109/TMI.2017.2723871}.

\bibitem[Sch{\"o}lkopf and Smola(2002)]{scholkopf}
Sch{\"o}lkopf B and Smola AJ.
\newblock {Learning with kernels: support vector machines, regularization,
  optimization, and beyond}.
\newblock MIT press, 2002.

\bibitem[Ongie et~al.(2017{\natexlab{a}})Ongie, Biswas, and Jacob]{gregtsp}
Ongie G, Biswas S, and Jacob M.
\newblock Convex recovery of continuous domain piecewise constant images from
  nonuniform fourier samples.
\newblock {IEEE Transactions on Signal Processing}, 2017{\natexlab{a}};
  66:\penalty0 236--250.

\bibitem[Ongie and Jacob(2016)]{gregpapers}
Ongie G and Jacob M.
\newblock Off-the-grid recovery of piecewise constant images from few fourier
  samples.
\newblock {SIAM Journal on Imaging Sciences}, 2016; 9:\penalty0 1004--1041.

\bibitem[Ongie and Jacob(2017)]{ongie2017fast}
Ongie G and Jacob M.
\newblock A fast algorithm for convolutional structured low-rank matrix
  recovery.
\newblock {IEEE Transactions on Computational Imaging}, 2017.

\bibitem[Ongie et~al.(2017{\natexlab{b}})Ongie, Willett, Nowak, and
  Balzano]{gregvariety}
Ongie G, Willett R, Nowak RD, and Balzano L.
\newblock Algebraic variety models for high-rank matrix completion.
\newblock In Precup D and Teh YW, editors, {Proceedings of the 34th
  International Conference on Machine Learning}, 06--11 Aug2017{\natexlab{b}};
  volume~70 of {Proceedings of Machine Learning Research}; pages 2691--2700,
  International Convention Centre, Sydney, Australia, 06--11
  Aug2017{\natexlab{b}}. PMLR.
\newblock URL \url{http://proceedings.mlr.press/v70/ongie17a.html}.

\bibitem[Roweis and Saul(2000)]{lle}
Roweis ST and Saul LK.
\newblock Nonlinear dimensionality reduction by locally linear embedding.
\newblock {science}, 2000; 290:\penalty0 2323--2326.

\bibitem[Belkin and Niyogi(2002)]{lapEig}
Belkin M and Niyogi P.
\newblock Laplacian eigenmaps and spectral techniques for embedding and
  clustering.
\newblock In {Advances in neural information processing systems}, 2002; pages
  585--591.

\bibitem[Lingala et~al.(2011)Lingala, Hu, DiBella, and Jacob]{ktslr}
Lingala SG, Hu~Y, DiBella E, and Jacob M.
\newblock Accelerated dynamic mri exploiting sparsity and low-rank structure:
  kt slr.
\newblock {IEEE transactions on medical imaging}, 2011; 30:\penalty0
  1042--1054.

\bibitem[Daubechies et~al.(2010)Daubechies, DeVore, Fornasier, and
  G{\"u}nt{\"u}rk]{irls}
Daubechies I, DeVore R, Fornasier M, and G{\"u}nt{\"u}rk CS.
\newblock Iteratively reweighted least squares minimization for sparse
  recovery.
\newblock {Communications on pure and applied mathematics}, 2010; 63:\penalty0
  1--38.

\bibitem[Zhao et~al.(2010)Zhao, Haldar, and Liang]{zhao_psf_2010}
Zhao B, Haldar JP, and Liang ZP.
\newblock {PSF} {Model}-{Based} {Reconstruction} with {Sparsity} {Constraint}:
  {Algorithm} and {Application} to {Real}-{Time} {Cardiac} {MRI}.
\newblock {Annual International Conference of the IEEE Engineering in Medicine
  and Biology Society.}, 2010; 2010:\penalty0 3390--3393.

\bibitem[Lingala and Jacob(2013)]{bcs}
Lingala SG and Jacob M.
\newblock Blind compressive sensing dynamic mri.
\newblock {IEEE transactions on medical imaging}, 2013; 32:\penalty0
  1132--1145.

\end{thebibliography}

\newpage
\begin{figure*}[t!]
\center
\includegraphics[scale=0.55]{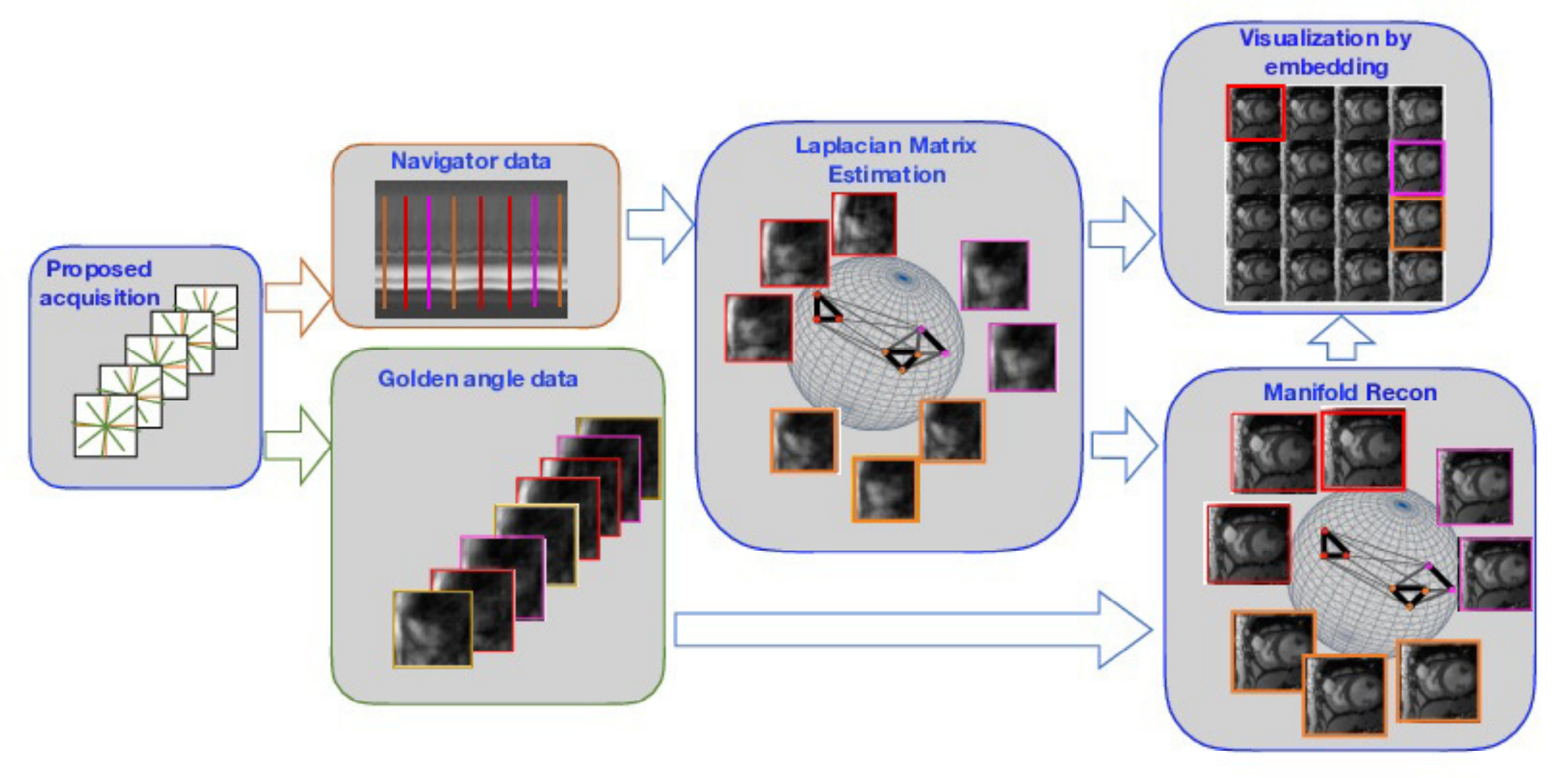}\\
\caption{\small Outline of the b-SToRM scheme. The free breathing and ungated data is acquired using a navigated golden angle acquisition scheme. We estimate the Laplacian matrix from the navigator data using the kernel low-rank model. The entries of the Laplacian matrix specifies the connectivity of the points of the manifold, with larger weights between similar frames in the dataset. The manifold is illustrated by the sphere, while the connectivity of the points are denoted by lines whose thickness is indicative of the proximity on the manifold. Note that the frames that may be closer on the manifold may be well separated in time. The bandlimited manifold recovery scheme uses the Laplacian matrix to recover the images from the acquired k-space measurements. The Laplacian matrix also facilitates the easy visualization of the data.}
\end{figure*}
\clearpage
\newpage
\begin{figure*}[t!]
\centering
\includegraphics[scale=0.7]{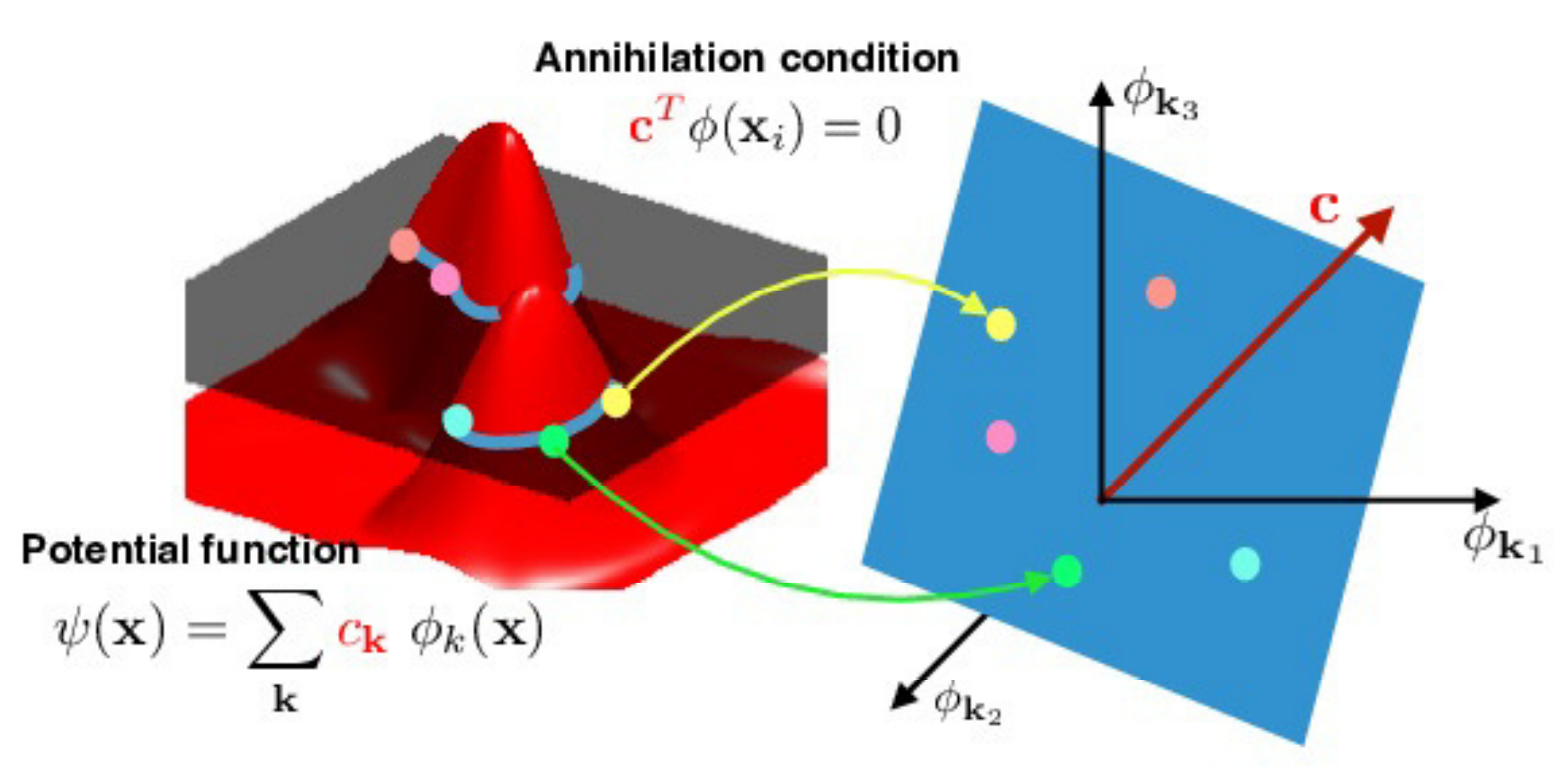}\\
\caption {\small Illustration of Annihilation Condition: The data points $\mathbf x$ lie on the zero-level set of a band-limited function $\psi$. Thus, each point $\mathbf x$ satisfies the relation: $\psi(\mathbf x) = 0$. The Fourier series co-efficients $\mathbf c$ satisfy the annihilation relation $\mathbf c^T \phi(\mathbf x) = 0$, where $\phi(\mathbf x)$ is a non-linear feature mapping of $\mathbf x$.}
\end{figure*}
\clearpage
\newpage
\begin{figure*}[t!]
\centering
\includegraphics[scale=0.2]{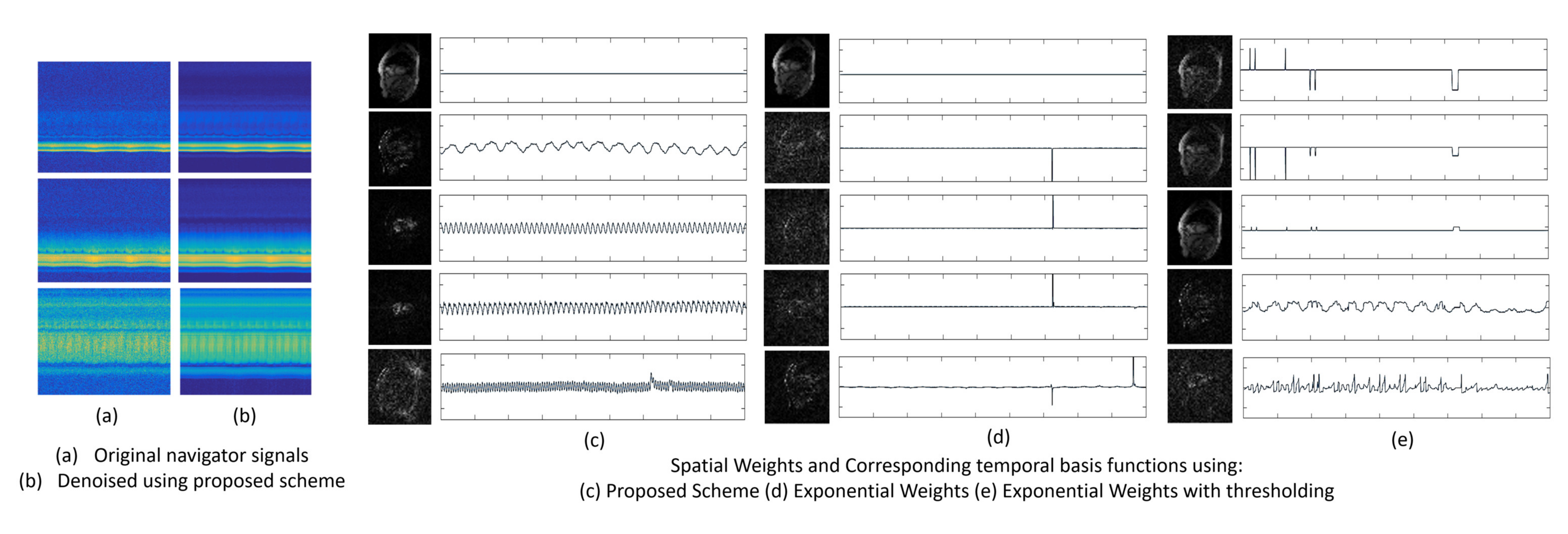}\\
\caption{\small Improved manifold Laplacian estimation using iterative low-rank approach: We compare the proposed scheme against the old SToRM approach that relies on exponential kernels. The proposed scheme denoises the original navigator signals in (a) using the low-rank approach to obtain (b). It is seen that the denoising approach significantly reduces the noise, while retaining the cardiac and respiratory motion information. The eigen functions $\mathbf V$ with the smallest eigen values of  the Laplacian estimated using the proposed scheme, exponential weights, and exponential weights with truncation (SToRM approach) are shown in (c), (d), and (e), respectively. It is observed that the proposed scheme provides more accurate estimates of cardiac and respiratory motion  than the other schemes, facilitating the low-rank approximation. Both (d) and (e) are affected by subtle motion of the subject while the proposed scheme is relatively unperturbed. The spatial coefficients $\mathbf U$ estimated for each case is also shown.}
\end{figure*}

\newpage
\begin{figure*}
\center
\includegraphics[scale=0.27]{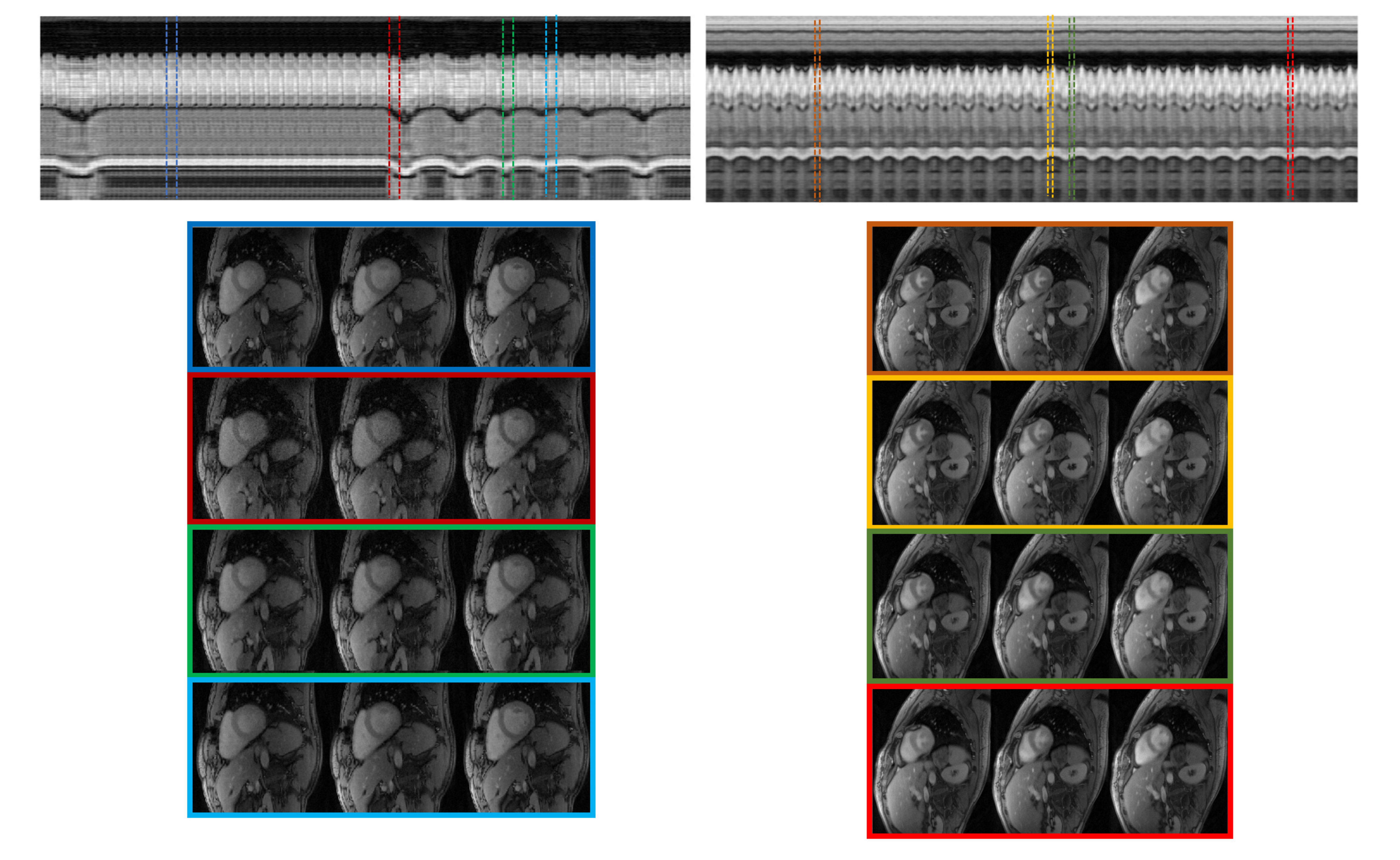}\\
\caption {\small Sensitivity of the algorithm to high motion: We illustrate the proposed scheme on two datasets acquired from two patients with different types of motion. For both datasets, we show a temporal profile for the whole acquisition to give an idea of the amount of breathing and cardiac motion present. We also show a few frames from time points with varying respiratory phase. The dataset on the left has regions with abrupt breathing motion at a few time points. Since these image frames have few similar frames in the dataset (poorly sampled neighbourhood on the manifold), the algorithm results in slightly noisy reconstructions at the time points with high breathing  motion (red box). The regions with low respiratory motion (blue and light blue boxes) are recovered well. The dataset on the right shows consistent, but low respiratory motion. By contrast, the heart rate in this patient was high. We observe that the proposed algorithm is able to produce good quality reconstructions in this case, since all neighbourhoods of the manifold are well sampled.}
\end{figure*}

\newpage
\begin{figure*}
\center
\includegraphics[scale=0.3]{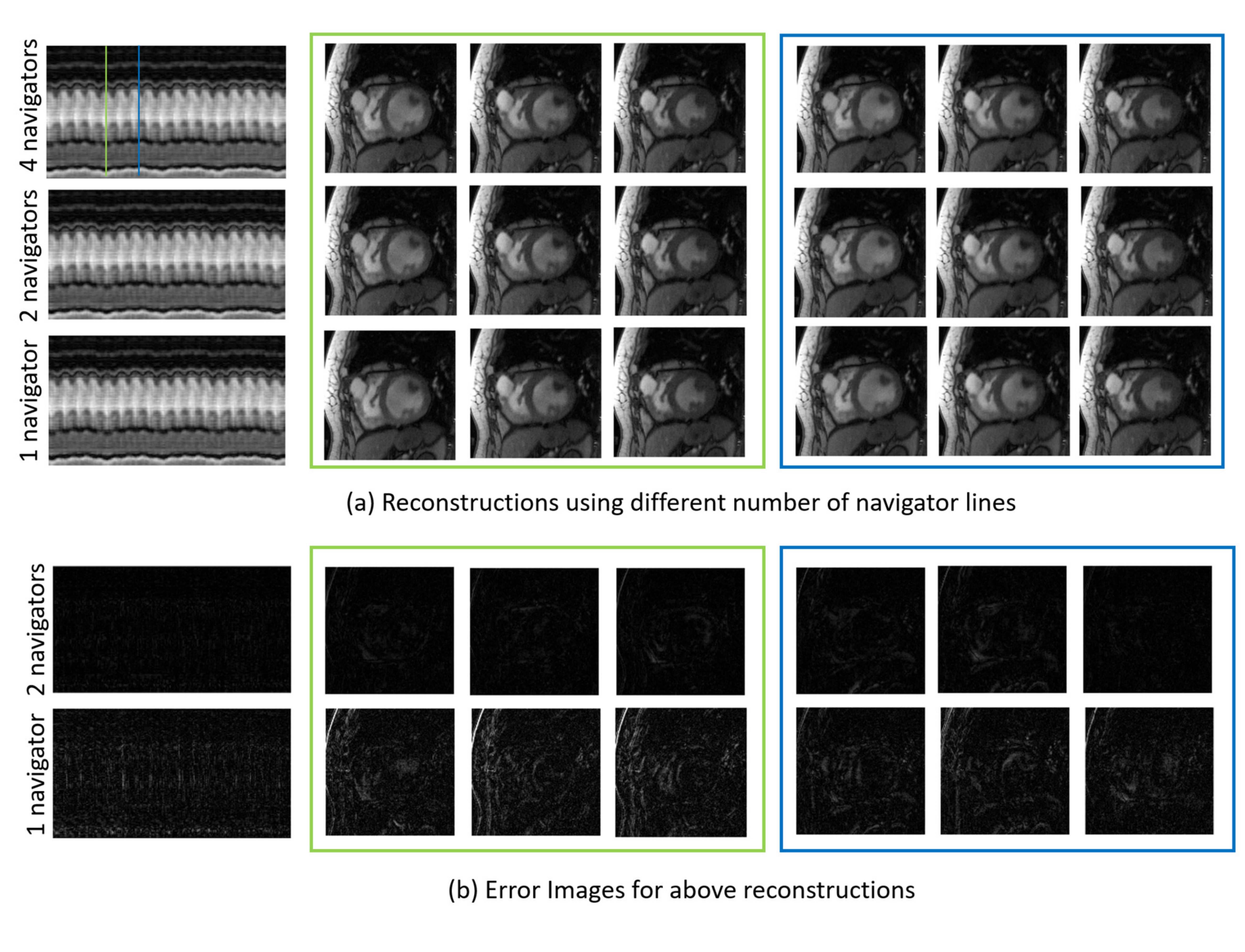}\\
\caption {\small Effect of number of navigator lines on the reconstruction quality. We perform an experiment to study the effect of computing the Laplacian matrix $\mathbf L$ from different number of navigator lines. For this purpose, we use one of the acquired datasets with 4 navigator lines per frame. We compute the ground-truth $\mathbf L$ matrix using all 4 navigators. Next, we also estimate the $\mathbf L$ matrix using 2 navigator lines (keeping only the $0^\circ$ and $90^\circ$ lines) and 1 navigator line (keeping only the $0^\circ$ line). We now reconstruct the full data using these three Laplacian matrices, as shown in the figure. We observe that two navigator lines are sufficient to compute the Laplacian matrix reliably. Using one navigator line induces some errors, especially in the frames highlighted in green which are from a time point with higher respiratory motion. As a comparison, note that the error images are in the same scale as those for Fig 6.}
\end{figure*}

\newpage
\begin{figure*}
\center
\includegraphics[scale=0.35]{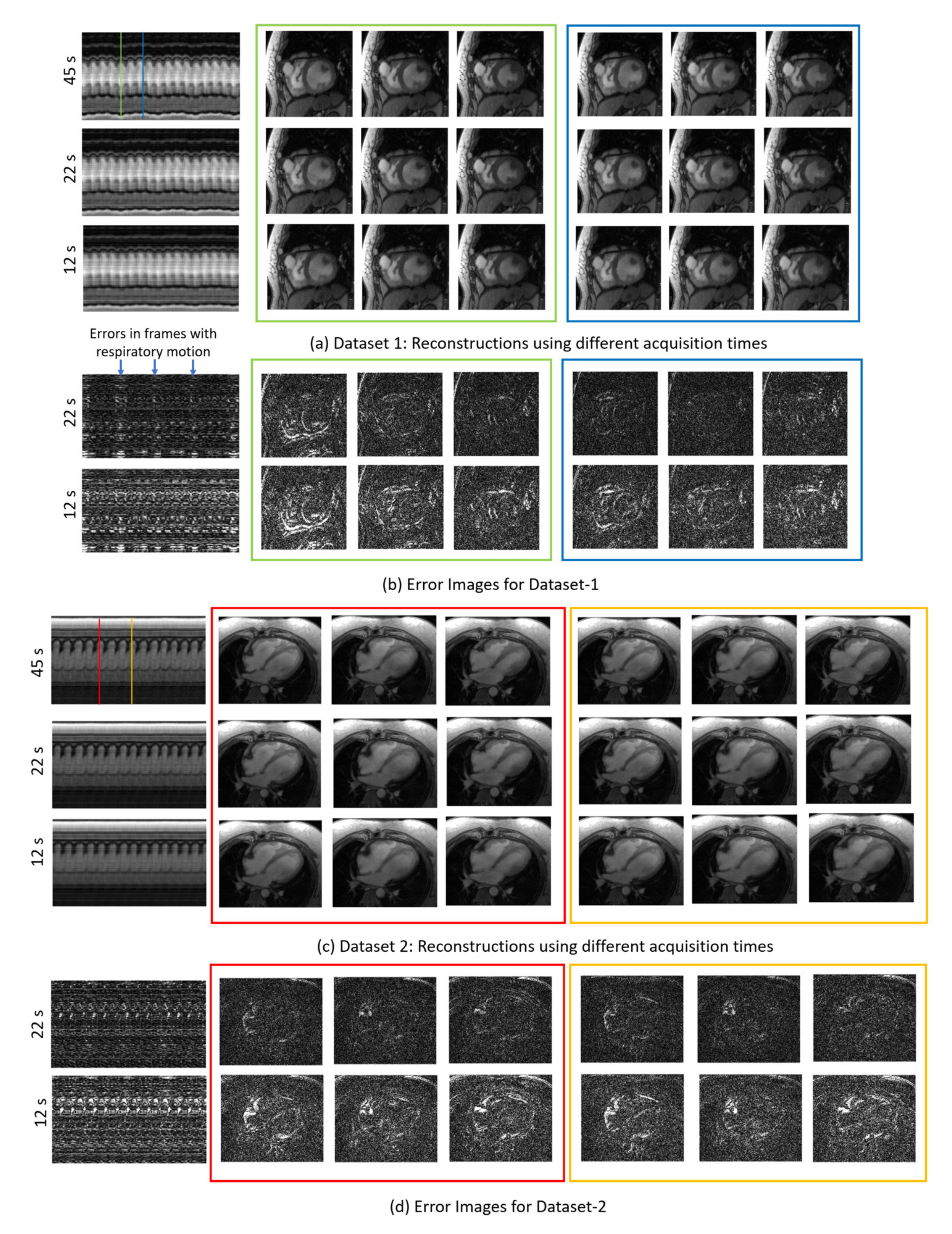}\\
\caption {\small Effect of number of frames on the reconstruction quality. We perform an experiment to study the effect of reconstructing the data from a fraction of the time-frames acquired. The original acquisition was 45 seconds long, resulting in 1000 frames. We compare the reconstruction of the $1^{st}$ 250 frames, using (1) all 1000 frames (2) only 550 frames, i.e. 22 s of acquisition (3) only 350 frames, i.e. 12 s of acquisition. As can be seen from the temporal profiles, Dataset-1 has more respiratory motion than Dataset-2. Consequently, the performance degradation in Dataset-1 is more pronounced with decrease in the number of frames. Moreover, the errors due to decrease in the number of frames is mostly seen in frames with higher respiratory motion, as pointed out by the arrows. As a comparison, note that the error images are in the same scale as those for Fig 5.}
\end{figure*}

\newpage
\begin{figure*}
\center
\includegraphics[scale=0.5]{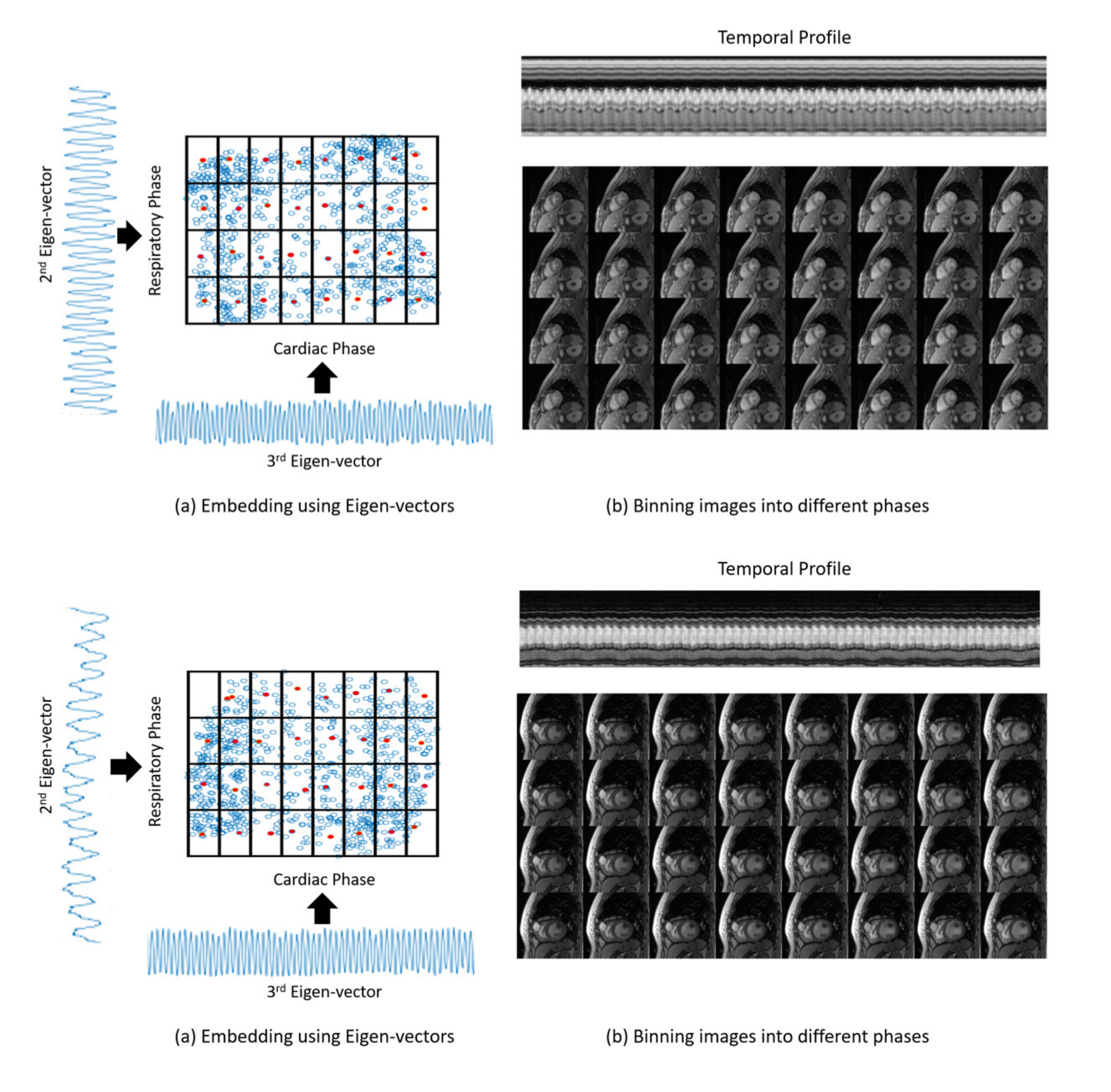}\\
\caption {\small Binning into cardiac and respiratory phases. We demonstrate that the reconstructed ungated image series can easily be converted to a gated series of images if desired. For this purpose, the $2^{nd}$ and $3^{rd}$ eigen-vectors of the estimated Laplacian matrix are used as an estimate of the respiratory and cardiac phases respectively. The images can then be separated into the desired number of cardiac and respiratory bins. Here, we demonstrate this on two datasets that have been separated into 8 cardiac and 4 respiratory phases. Representative images from these bins have been shown in the figure.}
\end{figure*}

\newpage
\begin{figure*}
\center
\includegraphics[scale=0.22]{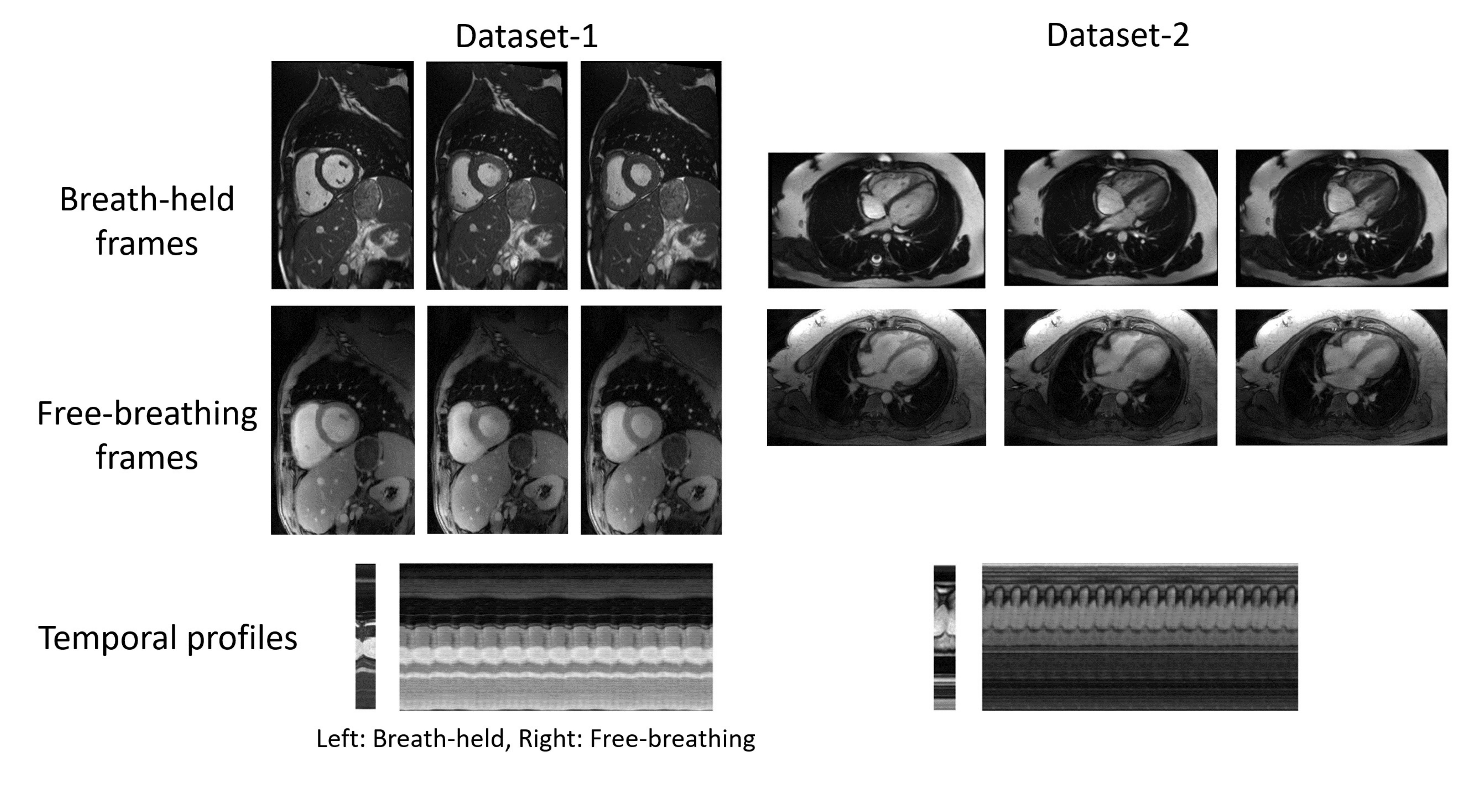}\\
\caption {\small Comparison to breath-held scheme. We demonstrate that our proposed free-breathing reconstruction technique produces images of similar quality to clinical breath-held scans, in the same acquisition time. Note that there are differences between the free-breathing and breath-held images due to variations in contrast between TRUFI and FLASH acquisitions, and also due to mismatch in slice position. However, the images we obtain are of clinically acceptable quality. Moreover, unlike the breath-held scheme we reconstruct the whole image time series (as is evident from the temporal profile). This can provide richer information , such as studying the interplay of cardiac and respiratory motion.}
\end{figure*}

\end{document}